%% file: neurips_2026.tex
\documentclass{article}

\PassOptionsToPackage{numbers,sort&compress}{natbib}
\usepackage[preprint]{neurips_2026} 

\usepackage[utf8]{inputenc}  
\usepackage[T1]{fontenc}     
\usepackage{hyperref}        
\usepackage{url}             
\usepackage{booktabs}        
\usepackage{amsfonts}        
\usepackage{nicefrac}        
\usepackage{microtype}       
\usepackage{xcolor}          
\usepackage{colortbl}        
\usepackage{multirow}        
\usepackage{enumitem}        
\usepackage{graphicx}        
\usepackage{caption}         
\graphicspath{{figures/}}
\usepackage[ruled]{algorithm} 
\usepackage{algpseudocode}   
\usepackage{float}
\usepackage{amsmath}
\usepackage{makecell}
\usepackage{pifont}
\newcommand{\cmarkG}{\textcolor{green!50!black}{\ding{51}}}
\newcommand{\xmarkR}{\textcolor{red!70!black}{\ding{55}}}
\newcommand{\notcheckmark}{\textcolor{orange!85!black}{\ensuremath{\triangle}}}
\usepackage[table]{xcolor}
\definecolor{HeaderBlue}{HTML}{EAF1FB}
\definecolor{SubHeaderBlue}{HTML}{F3F6FA}
\definecolor{RowGray}{HTML}{F7F7F7}
\definecolor{StrongRow}{HTML}{FFF4E6}

\title{AgentEscapeBench: Evaluating Out-of-Domain Tool-Grounded Reasoning in LLM Agents}

\author{
Zhengkang Guo\textsuperscript{1,2\,*}\,,\quad
Yiyang Li\textsuperscript{2\,*}\,,\quad
Lin Qiu\textsuperscript{2\,\dag}\,,\quad
Xiaohua Wang\textsuperscript{2}\,,\quad
Jingwen Xv\textsuperscript{2}\,,\quad
Dongyu Ru\textsuperscript{2}\,,\\
\bfseries Xiaoyu Li\textsuperscript{2}\,,\quad
Xiaoqing Zheng\textsuperscript{1\,\ddag}\,,\quad
Xuezhi Cao\textsuperscript{2}\,,\quad
Xunliang Cai\textsuperscript{2}\\[0.4em]
\textsuperscript{1}Fudan University\qquad
\textsuperscript{2}Meituan Longcat Team\\[0.3em]
\texttt{zkguo24@m.fudan.edu.cn},\quad
\texttt{\{liyiyang06,\,qiulin07\}@meituan.com}
}

\begin{document}

\maketitle

\renewcommand{\thefootnote}{\fnsymbol{footnote}}
\footnotetext[1]{Equal contribution.}
\footnotetext[2]{Project leader.}
\footnotetext[3]{Corresponding author.}
\renewcommand{\thefootnote}{\arabic{footnote}}

\begin{abstract}
\input{abstract_new.tex}
\end{abstract}

\input{introduction_new.tex}
\input{data_construction2.tex} 

\input{main_experiment.tex}

\input{fine_grained_analysis2.tex}

\input{related_work.tex}
\input{conclusion.tex}

\begin{ack}
\end{ack}

{
\small
\bibliographystyle{plainnat}
\bibliography{references}
}

\appendix
\input{appendix_data_construction.tex}
\input{appendix_evaluation.tex}
\input{appendix_trajectory.tex}
\input{appendix_limitations.tex}
\newpage

\end{document}

%% file: abstract_new.tex
As LLM-based agents increasingly rely on external tools, it is important to evaluate their ability to sustain tool-grounded reasoning beyond familiar workflows and short-range interactions. We introduce \textsc{AgentEscapeBench}, an escape-room-style benchmark that tests whether agents can infer, execute, and revise novel tool-use procedures under explicit long-range dependency constraints. Each task defines a directed acyclic dependency graph over tools and items, requiring agents to invoke real external functions, track hidden state revealed incrementally, propagate intermediate results, and submit a deterministically verifiable final answer.
AgentEscapeBench includes 270 instances across five difficulty tiers and supports fully automated evaluation. Experiments with sixteen LLM agents and human participants show that performance drops sharply as dependency depth increases: humans decline from 98.3\% success at difficulty-5 to 80.0\% at difficulty-25, while the best model drops from 90.0\% to 60.0\%. Trajectory analysis attributes model failures mainly to breakdowns in long-range state tracking, clue adherence, and intermediate-result propagation. These findings suggest that current agents can often handle local tool use but still struggle with deep contextual dependencies. We hope AgentEscapeBench can serve as a diagnostic testbed for measuring current agent capabilities and informing future training efforts toward more robust general-purpose reasoning, action, and adaptation.

%% file: introduction_new.tex
\section{Introduction}

\label{sec:intro}

Tool-calling has expanded the role of large language models (LLMs) beyond conversational question answering, enabling them to invoke APIs, execute code, manipulate files, and participate in real-world workflows. This shift has led to a new class of LLM-based agents with substantial practical potential~\citep{claudecode}. As such agents are increasingly deployed in production settings, rigorous and diagnostic evaluation of their capabilities becomes essential.

Existing agent benchmarks evaluate important aspects of this capability. Tool-calling benchmarks such as BFCL~\citep{patil2025berkeley} focus on individual API invocations, including tool selection, schema following, and argument generation. More interactive benchmarks, including Tau$^2$-Bench~\citep{barres2025tau}, TripBench~\citep{shen2026trip}, VitaBench~\citep{he2025vitabench}, SWE-bench~\citep{jimenez2023swe}, and Gaia2~\citep{russell2025gaia}, extend evaluation to multi-step tasks in domains such as travel planning, customer service, software engineering, mobile applications, and web workflows. However, many of these tasks are grounded in familiar domains with recurring solution templates: booking travel, applying customer-service policies, or editing code and running tests. Strong performance may therefore reflect learned domain conventions or short-range reactive tool use, rather than the ability to infer and execute a novel tool-use procedure in an unfamiliar environment.

Puzzle-based and escape-room-style benchmarks offer a complementary way to reduce reliance on familiar domain priors. Benchmarks such as EscapeBench~\citep{qian2024escapebench} and PuzzleWorld~\citep{li2025puzzleworld} probe exploratory reasoning in unfamiliar settings, but many operate in text-only action spaces or question-answering formats. As a result, they do not fully capture modern tool-grounded deployments, where agents must call external functions with valid parameters, interpret structured feedback, and propagate outputs through executable dependency chains.

A further limitation is that long-horizon interaction does not necessarily require deep contextual dependency. In many benchmarks, tool calls can often be selected from the most recent observation, while information introduced much earlier is rarely an explicit prerequisite for later actions. This makes it difficult to diagnose whether failures stem from forgotten clues, incorrect state tracking, faulty propagation of intermediate results, or poor tool selection. A more diagnostic benchmark should therefore make long-context dependencies explicit, traceable, and automatically analysable. Table~\ref{table:related_work} summarises these distinctions.

\begin{figure}[t]
  \centering
  \includegraphics[width=\linewidth]{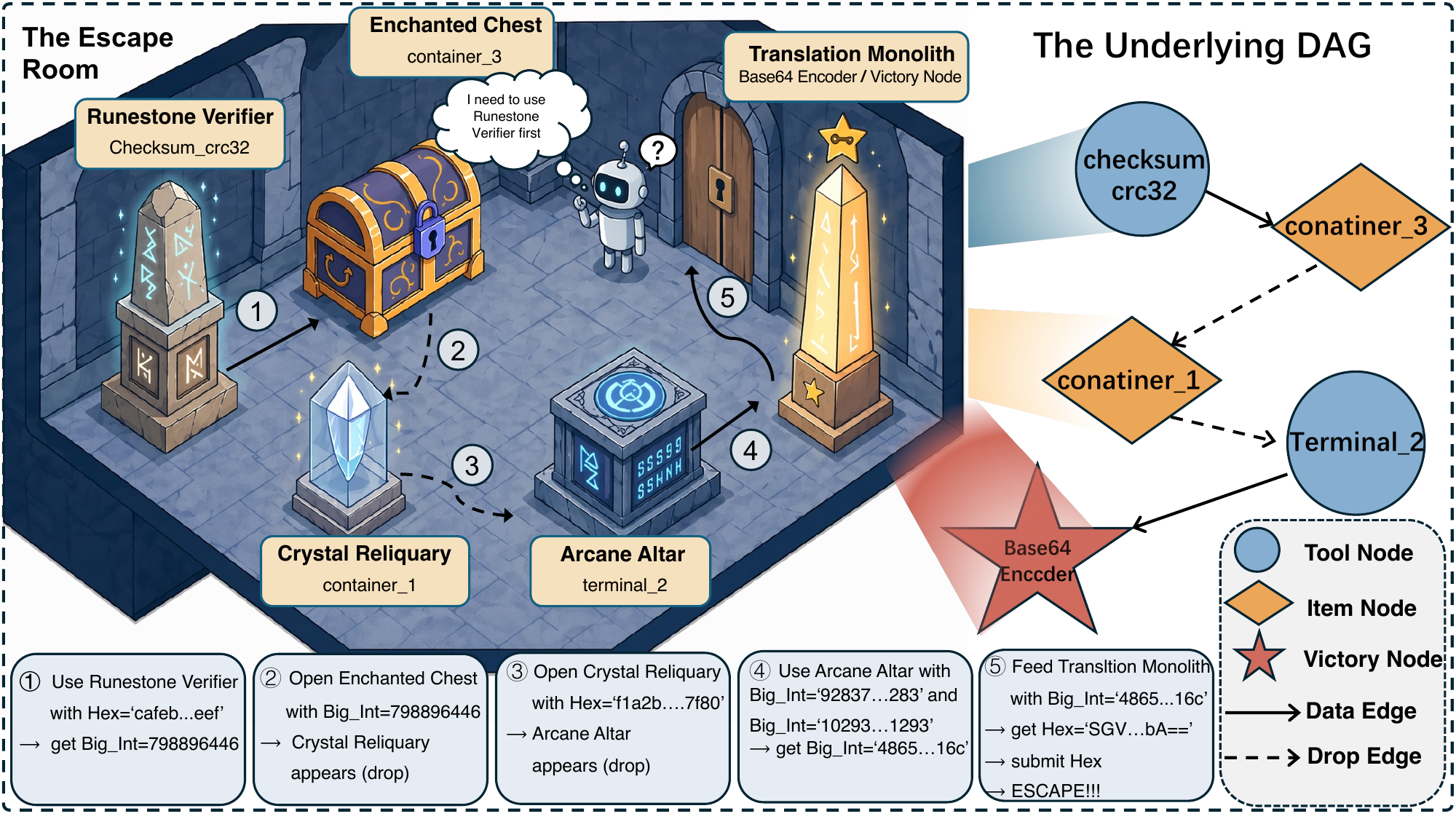}
  \caption{%
    \textbf{Conceptual illustration of AgentEscapeBench.}
    The agent is placed in a themed escape room populated with unfamiliar tools and hidden items.
    It must explore the environment, invoke tools with correct parameters derived from narrative clues, and propagate intermediate outputs through a multi-step dependency chain to unlock the final exit.
  }
  \label{fig:concept}
\end{figure}

To address these gaps, we introduce \textbf{AgentEscapeBench}, an escape-room-style framework for evaluating tool-grounded agent reasoning in unfamiliar environments; Figure~\ref{fig:concept} provides a conceptual overview. Each task is designed so that the agent cannot rely on domain-specific prior knowledge or a fixed workflow template. Instead, it must extract clues from the task description and incrementally revealed observations, decide which tools to call and how to parameterise them, propagate intermediate outputs through dependency chains, and revise its plan as new information becomes available. This design targets the underlying reasoning and tool-use capabilities needed for general-purpose agents, rather than surface-level mastery of a particular workflow.

We instantiate AgentEscapeBench with 270 tasks across five difficulty tiers, defined by DAGs with 5, 10, 15, 20, and 25 nodes. Each task is generated from compositional tool and item templates, supports incremental information disclosure, executes real external functions, and has a deterministic ground-truth answer for automated evaluation. We evaluate sixteen representative LLMs and conduct a human study under the same interaction and feedback conditions. Our results show that AgentEscapeBench differentiates models across a wide capability spectrum: performance drops sharply as dependency depth increases, and trajectory-level analyses reveal model-specific failure signatures in behaviour, tool use, and problem-solving efficiency. Our main contributions are as follows:

\begin{itemize}[leftmargin=1.5em]
  \item \textbf{Out-of-domain, tool-grounded agent evaluation.}
  We formulate AgentEscapeBench as an escape-room-style benchmark that removes familiar domain priors and fixed workflow templates, thereby targeting transferable tool-grounded reasoning.

  \item \textbf{Controllable DAG-based task generation.}
  We introduce an automated pipeline that composes tool and item templates into diverse DAG-structured tasks with incremental information disclosure, executable tool calls, and deterministic ground-truth answers.

  \item \textbf{Model and human evaluation across difficulty tiers.}
  We construct 270 instances across five controlled difficulty levels and evaluate sixteen LLMs alongside human participants, revealing how performance degrades with increasing dependency depth.

  \item \textbf{Trajectory- and step-level diagnostic analysis.}
  We analyse behavioural patterns, tool-calling errors, and problem-solving efficiency, exposing model-specific failure signatures that diagnose current capability bottlenecks and provide actionable signals for improving general-purpose agent capabilities.
\end{itemize}

%% file: data_construction2.tex
\section{Dataset Construction}
\label{sec:data_construction}


AgentEscapeBench instances are produced by a six-stage automated pipeline (Figure~\ref{fig:pipeline}).
Each instance is a directed acyclic graph (DAG) of tool and item nodes, accompanied by natural-language narratives that serve as inferential clues.
We provide a concise overview here; full algorithmic details appear in Appendix~\ref{app:data_construction}.

{%
\setlength{\parskip}{0.3em}%

\noindent\textbf{Stage 1: Template library.}\quad
We curate a library of \textbf{32 tool templates}, \textbf{2 terminal templates}, \textbf{16 item templates}, and \textbf{4 container templates}.
Each tool template defines typed input ports, typed output ports, and a deterministic execution function.
Tools span big-integer arithmetic, cryptographic primitives, encoding/decoding, file operations, and graph algorithms, among others (full list in Appendix~\ref{app:data_construction}).
Item and container templates carry information payloads and support an \emph{incremental-disclosure mechanism}: when a node whose output ports include \texttt{ITEM} or \texttt{HIDDEN\_ITEM} types is successfully resolved, the connected downstream nodes become newly visible to the agent, revealing items or tools that were not accessible at the outset.

\noindent\textbf{Stage 2: DAG skeleton generation.}\quad
Given a target node count $n$ (the difficulty level), we assemble a DAG skeleton using a \textbf{reverse-generation algorithm} (see Appendix~\ref{app:data_construction} for full details).
Starting from a randomly sampled goal node, the algorithm iteratively resolves pending input-port requirements by either reusing an existing node whose output type matches, or instantiating a fresh template.
Structural constraints---acyclicity, single-use semantics for physical-object ports, and exclusivity for semantically sensitive ports---are enforced at every edge creation.
Isomorphic duplicates are discarded.

\begin{figure}[t]
  \centering
  \includegraphics[width=\linewidth]{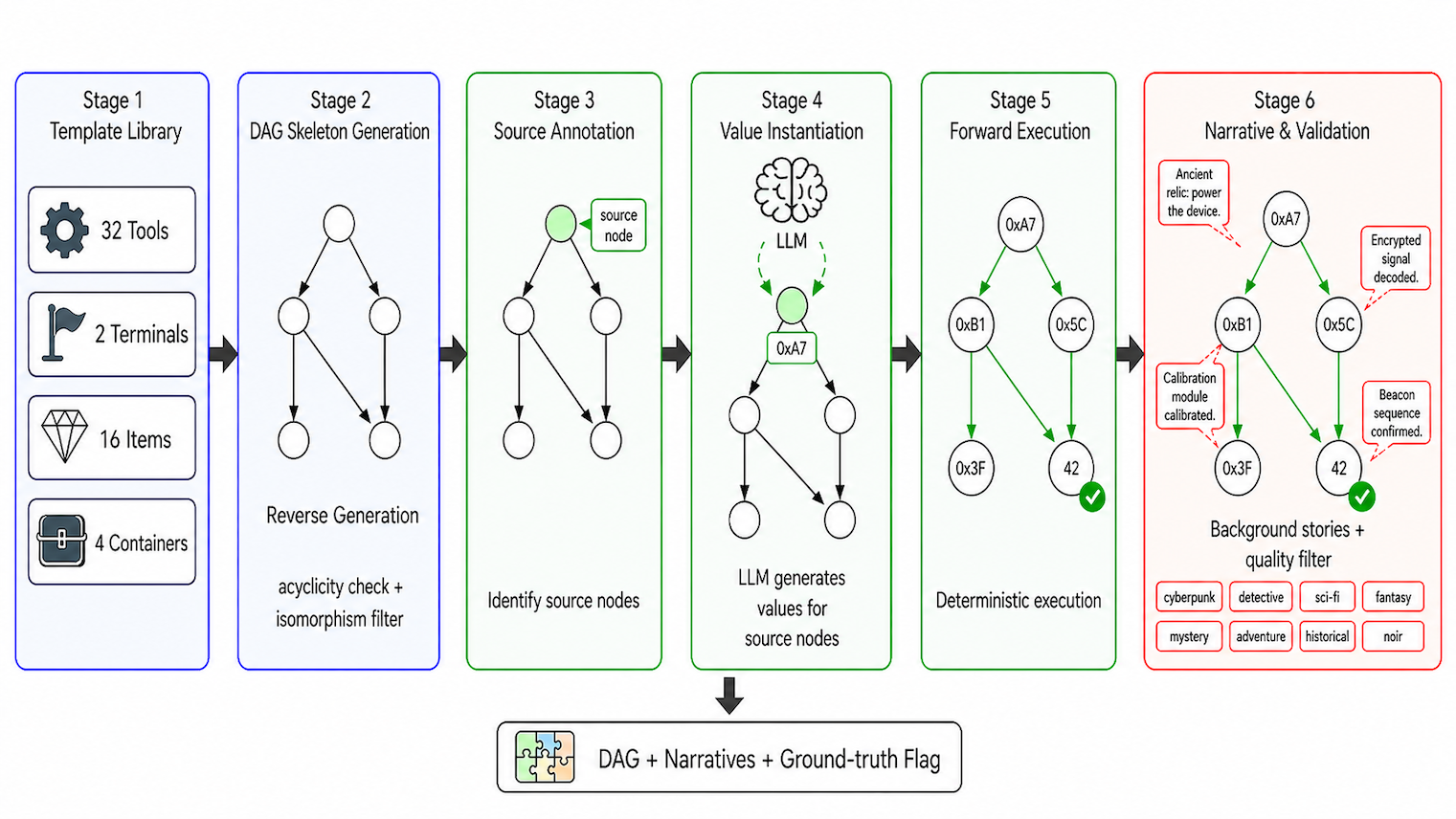}
  \caption{%
    \textbf{Six-stage data construction pipeline of AgentEscapeBench.}
    Starting from a curated template library (Stage~1), a reverse-generation algorithm assembles a DAG skeleton (Stage~2), annotates source ports (Stage~3), instantiates concrete values via an LLM (Stage~4), executes the DAG forward to compute ground-truth outputs (Stage~5), and generates themed narratives with structural validation (Stage~6).
  }
  \label{fig:pipeline}
\end{figure}

\noindent\textbf{Stages 3--4: Source annotation and value instantiation.}\quad
All leaf-level input ports (those with no incoming edge) are identified as sources requiring concrete seed values.
An LLM generates these values, respecting type contracts and inter-port consistency constraints.

\noindent\textbf{Stage 5: Deterministic forward execution.}\quad
The DAG is executed in topological order: each tool reads its inputs, applies its deterministic function, and propagates results along outgoing edges.
The terminal output constitutes the unique ground-truth flag.
Instances that fail the forward pass are discarded.

\noindent\textbf{Stage 6: Narrative generation and validation.}\quad
An LLM generates background narratives for each node, partially disclosing the data-flow structure.
Each instance is assigned one of eight thematic styles to promote diversity.
A final automated quality check verifies that the instance is fully instantiated and solvable; only instances that pass are admitted to the benchmark.

\noindent\textbf{Human solvability validation.}\quad
Beyond automated checks, we conduct a human study to provide the strongest guarantee of data quality.
All instances are distributed across multiple participants, with each instance attempted by at least one participant under the same interface and feedback conditions as the LLM agents.
Every instance in the benchmark is successfully solved by at least one human participant, confirming that all automatically generated puzzles are well-formed and solvable.

}%

%% file: main_experiment.tex
\section{Experiments}
\label{sec:experiments}

\definecolor{humanrow}{RGB}{232,245,233}
\definecolor{tablestripe}{RGB}{245,247,252}

\subsection{Evaluation Protocol}

AgentEscapeBench contains 60 puzzle instances at each of difficulty levels 5, 10, 15, and 20, and 30 instances at difficulty 25, yielding 270 instances in total.
Each evaluation episode proceeds as a multi-turn dialogue between the agent and a sandbox environment.
At the start of an episode, the agent receives an initial observation that describes the room setting, the available tools and items, and the background narratives associated with visible nodes.
On every subsequent turn, the agent may either \emph{inspect} an entity to retrieve its detailed description, or \emph{invoke} a tool by specifying the tool name and parameter values.
The environment returns execution results for correct invocations or structured error feedback (e.g., wrong parameter value, missing dependency) for incorrect ones.
When a successful tool invocation triggers a drop event, newly revealed items and tools become visible in the next observation, realising the incremental-disclosure mechanism described in Section~\ref{sec:data_construction}.
A complete solved trajectory illustrating this interaction loop is provided in Appendix~\ref{app:trajectory}.
An episode is scored as a \emph{success} if and only if the agent produces the correct output value for the terminal (victory) node and submits it to the environment.
The ground-truth value is deterministically computed during data construction (Section~\ref{sec:data_construction}, Stage~5), ensuring unambiguous, fully automated evaluation with no human judgement required.
We evaluate sixteen models in total; step budgets, model list, and hyperparameters are detailed in Appendix~\ref{app:evaluation}.
To obtain a human performance baseline, we randomly assign instances across multiple participants such that each instance is attempted exactly once by one participant, and report the aggregated success rate as the human baseline in Table~\ref{tab:main}.

\begin{table}[t]
  \centering
  \caption{%
    \textbf{Main results on AgentEscapeBench.}
    For each difficulty level (DAG node count), we report success rate (\textbf{SR}, \%), sub-problem resolution rate (\textbf{Sub.}, \%), and hidden-node discovery rate (\textbf{Disc.}, \%).
    \textbf{Bold} = best model in column; \underline{underline} = second best.
    $\Delta_{\text{SR}}$: absolute SR drop from difficulty-5 to the hardest evaluated level.
    The human baseline provides a reference for achievable performance under the same evaluation conditions.
  }
  \label{tab:main}
  \resizebox{\linewidth}{!}{%
  \begin{tabular}{l ccc ccc ccc ccc ccc c}
    \toprule
    & \multicolumn{3}{c}{\textbf{Difficulty-5}} & \multicolumn{3}{c}{\textbf{Difficulty-10}} & \multicolumn{3}{c}{\textbf{Difficulty-15}} & \multicolumn{3}{c}{\textbf{Difficulty-20}} & \multicolumn{3}{c}{\textbf{Difficulty-25}} & \\
    \cmidrule(lr){2-4} \cmidrule(lr){5-7} \cmidrule(lr){8-10} \cmidrule(lr){11-13} \cmidrule(lr){14-16}
    \textbf{Model} & \textbf{SR} & \textbf{Sub.} & \textbf{Disc.} & \textbf{SR} & \textbf{Sub.} & \textbf{Disc.} & \textbf{SR} & \textbf{Sub.} & \textbf{Disc.} & \textbf{SR} & \textbf{Sub.} & \textbf{Disc.} & \textbf{SR} & \textbf{Sub.} & \textbf{Disc.} & $\boldsymbol{\Delta_{\text{SR}}}$ \\
    \midrule
    \rowcolor{humanrow}
    Human$^\ddagger$     & 98.3 & 96 & 100.0 & 95.0 & 94 & 97.3 & 85.0 & 92 & 96.8 & 81.7 & 90 & 95.9 & 80.0 & 86 & 95.4 & 18.3 \\
    \midrule
    \rowcolor{tablestripe}
    Claude-Opus-4.6      & 90.0 & \underline{93} & \underline{99.1} & \textbf{83.3} & \underline{86} & \underline{93.9} & \underline{80.0} & \underline{86} & \textbf{96.3} & \textbf{71.0} & \textbf{86} & \textbf{96.5} & \textbf{60.0} & \textbf{82} & \textbf{94.9} & 30.0 \\
    Gemini-3.1-Pro-Preview & 91.7 & \textbf{94} & \textbf{100.0} & \textbf{83.3} & \textbf{89} & \textbf{94.5} & \textbf{83.3} & \textbf{89} & \underline{94.6} & \underline{60.0} & \underline{82} & \underline{86.6} & 13.3 & \underline{74} & \underline{74.7} & 78.4 \\
    \rowcolor{tablestripe}
    GPT-5.4              & \textbf{96.7} & 92 & \underline{99.1} & \textbf{83.3} & 83 & 90.9 & 70.0 & 78 & 88.6 & 53.3 & 67 & 75.2 & \underline{43.3} & 71 & 68.8 & 53.4 \\
    Kimi-K2.5            & \underline{95.0} & 92 & \textbf{100.0} & \underline{81.7} & \underline{86} & 91.1 & 70.0 & 78 & 81.5 & 31.7 & 66 & 80.1 & --- & --- & --- & 63.3 \\
    \rowcolor{tablestripe}
    DeepSeek-Chat        & 86.7 & 89 & 98.2 & 76.7 & 85 & 91.2 & 58.3 & 78 & 91.2 & 33.3 & 66 & 77.3 & --- & --- & --- & 53.4 \\
    DeepSeek-Reasoner    & 78.3 & 89 & \underline{99.1} & 65.0 & 81 & 89.9 & 58.3 & 77 & 85.8 & 36.7 & 68 & 80.4 & --- & --- & --- & 41.6 \\
    \rowcolor{tablestripe}
    Gemini-3-Flash-Preview & 88.3 & 93 & \textbf{100.0} & 66.7 & 84 & 90.5 & 50.0 & 75 & 83.6 & 18.3 & 62 & 76.0 & --- & --- & --- & 70.0 \\
    Doubao-Seed-2.0-Pro  & 86.7 & 89 & 98.5 & 63.3 & 76 & 85.8 & 31.7 & 59 & 70.9 & 15.0 & 49 & 70.9 & --- & --- & --- & 71.7 \\
    \rowcolor{tablestripe}
    GPT-5                & 56.7 & 67 & 78.9 & 48.3 & 66 & 67.6 & 40.0 & 61 & 69.3 & 20.0 & 25 & 27.7 & --- & --- & --- & 36.7 \\
    MiniMax-M2           & 70.0 & 88 & 98.2 & 51.7 & 75 & 85.8 & 28.3 & 59 & 75.7 & 5.0 & 43 & 56.2 & --- & --- & --- & 65.0 \\
    \rowcolor{tablestripe}
    Kimi-K2              & 51.7 & 55 & 62.1 & 35.0 & 48 & 53.3 & 23.3 & 43 & 55.0 & 11.7 & 32 & 47.8 & --- & --- & --- & 40.0 \\
    \midrule
    \rowcolor{tablestripe}
    Qwen3-235B-A22B$^\dagger$      & 88.3 & 89 & 99.1 & 58.3 & 75 & 87.0 & 16.7 & 50 & 63.9 & --- & --- & --- & --- & --- & --- & 71.6 \\
    Qwen3-Next-80B-A3B$^\dagger$   & 76.7 & 84 & 96.4 & 35.0 & 62 & 78.5 & 10.0 & 42 & 62.4 & --- & --- & --- & --- & --- & --- & 66.7 \\
    \rowcolor{tablestripe}
    Qwen3-32B$^\dagger$            & 60.0 & 75 & 89.1 & 16.7 & 48 & 57.5 & 10.0 & 32 & 48.3 & --- & --- & --- & --- & --- & --- & 50.0 \\
    Qwen3-14B$^\dagger$            & 50.0 & 65 & 84.4 & 11.7 & 42 & 61.1 &  6.7 & 19 & 46.8 & --- & --- & --- & --- & --- & --- & 43.3 \\
    \rowcolor{tablestripe}
    Qwen3-8B$^\dagger$             & 28.3 & 43 & 73.0 &  0.0 & 22 & 53.2 &  0.0 & 16 & 46.8 & --- & --- & --- & --- & --- & --- & 28.3 \\
    \bottomrule
    \multicolumn{17}{l}{\footnotesize $^\ddagger$Human participants attempt instances under the same interface and feedback conditions as LLM agents; instances are randomly distributed among participants (one attempt per instance).}
  \end{tabular}%
  }
\end{table}

\subsection{Main Results}

Table~\ref{tab:main} reports the performance of sixteen representative LLMs across five difficulty levels.
For each model--difficulty pair, we report three metrics: \textbf{success rate} (SR, the fraction of instances fully solved), \textbf{sub-problem resolution rate} (Sub., the average fraction of DAG nodes correctly resolved per instance), and \textbf{hidden-node discovery rate} (Disc., the average fraction of initially hidden nodes that the agent successfully uncovers during an episode).
Sub-problem resolution rate provides a more fine-grained view of partial progress than the binary success rate, while hidden-node discovery rate measures the agent's ability to explore beyond the initially visible environment by triggering the incremental-disclosure mechanism.
The first row reports human performance as a reference baseline; the upper model block contains eleven models evaluated on difficulty 5--20 (and three on difficulty-25); the lower block contains five open-source Qwen3 variants evaluated on difficulty 5--15 only (marked $^\dagger$).
Not all models are evaluated at every difficulty level: for weaker models whose scores are already very low at easier levels, the first few tiers suffice to reveal their capability profile and performance trend; stronger models are extended to higher difficulties for a more thorough assessment; difficulty-25 is reserved for the three top-performing models (Claude-Opus-4.6, GPT-5.4, and Gemini-3.1-Pro-Preview) to probe their behaviour under extreme task depth.
Within each model block, models are sorted by average success rate in descending order.

\subsection{Key observations.}
\label{sec:analysis}

\paragraph{Performance degradation across difficulty tiers.}
All three metrics degrade monotonically with difficulty for every model, with the steepest drops occurring between difficulty-15 and difficulty-20, where most models lose more than half their difficulty-5 success rate.
This degradation is not uniform: the $\Delta_{\text{SR}}$ column reveals that some models experience a catastrophic collapse (e.g., Gemini-3.1-Pro-Preview drops 78.4 percentage points from difficulty-5 to difficulty-25; Doubao-Seed-2.0-Pro drops 71.7 points), while Claude-Opus-4.6 exhibits the most graceful degradation ($\Delta = 30.0$ points) and retains the highest absolute performance at all difficulty levels above 5.
The divergence suggests that aggregate performance on easy tasks is an unreliable proxy for deep reasoning ability: a model can achieve high scores at difficulty-5 by exploiting shallow heuristics, but such strategies fail to scale to longer dependency chains where genuine multi-step planning is required.
Difficulty-25 further separates even the top-performing models: Claude-Opus-4.6 (60.0\%) leads, followed by GPT-5.4 (43.3\%) and Gemini-3.1-Pro-Preview (13.3\%), revealing qualitatively different scaling behaviours under extreme task depth.

\paragraph{Sub-problem resolution reveals the chaining bottleneck.}
Sub-problem resolution and discovery rates reveal meaningful partial progress even when end-to-end success rates approach zero.
At difficulty-20, MiniMax-M2 solves only 5.0\% of instances end-to-end but still resolves 43\% of sub-problems and discovers 56.2\% of hidden nodes; similarly, Gemini-3.1-Pro-Preview maintains 82\% sub-problem resolution and 86.6\% hidden-node discovery even as its success rate drops to 60.0\%.
This pattern indicates that the primary bottleneck lies not in individual tool calls but in \emph{chaining} intermediate results across long dependency paths---a capability that degrades superlinearly with graph depth.

\paragraph{Reasoning models do not dominate.}
Models with explicit chain-of-thought or reasoning capabilities do not uniformly outperform their non-reasoning counterparts.
DeepSeek-Reasoner (avg.\ SR 59.6\%) underperforms DeepSeek-Chat (avg.\ 63.8\%) across all difficulty levels, despite the former being specifically designed for extended deliberation.
This suggests that the reasoning bottleneck in AgentEscapeBench is not the \emph{depth} of inference within a single reasoning trace, but rather the ability to \emph{ground} reasoning in real tool interactions and to \emph{update} beliefs as new environmental information arrives.
Extended internal reasoning may, in fact, introduce overconfidence or distraction when the environment is unfamiliar and feedback must come from external tool calls rather than from prior knowledge.

\paragraph{Human--model gap reveals reasoning generalisation deficits.}
The aggregated human success rate degrades gracefully with difficulty (98.3\% at difficulty-5 $\to$ 80.0\% at difficulty-25, a drop of 18.3 points), whereas even the best model (Claude-Opus-4.6) drops from 90.0\% to 60.0\% (30.0 points), and most other models collapse far more steeply.
At difficulty-15, humans achieve 85.0\% while the best model reaches 83.3\%; by difficulty-20, the gap widens (81.7\% human vs.\ 71.0\% best model).
This divergence cannot be attributed to unfamiliarity with specific tools or domains, because humans face the same unfamiliar environment.
Humans adapt readily to unfamiliar problem settings and apply given rules reliably even as task complexity grows, exhibiting strong capability generalisation; LLMs, by contrast, degrade sharply once they depart from familiar task templates, and this transfer deficit is amplified at higher difficulties.
These findings indicate that current LLMs have not yet achieved the robust, transferable tool-grounded reasoning ability that would close the gap with human intelligence in genuinely novel settings.

%% file: fine_grained_analysis2.tex
\section{Fine-Grained Trajectory Analysis}
\label{sec:fine_grained}

To move beyond aggregate success rates and diagnose \emph{how} models succeed or fail, we perform a systematic offline analysis of every evaluation trajectory.
For each of the 240 instances per model (60 per difficulty level 5--20), we parse the full interaction log and extract behavioural signals along three complementary dimensions: behavioural pattern analysis (\S\ref{sec:behavioural}), tool-calling error analysis (\S\ref{sec:error_taxonomy}), and efficiency analysis (\S\ref{sec:efficiency}).
The analysis covers eleven non-Qwen models across difficulty levels 5--20.

\subsection{Behavioural Pattern Analysis}
\label{sec:behavioural}

\begin{figure}[t]
  \centering
  \includegraphics[width=\linewidth]{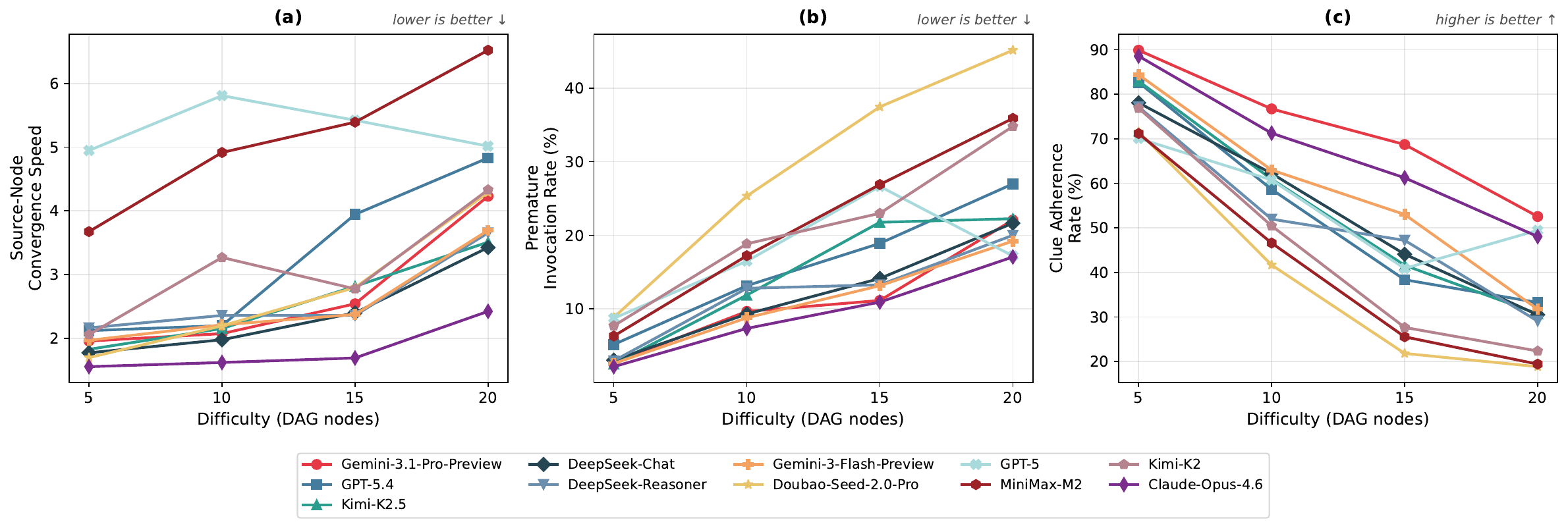}
  \caption{%
    \textbf{Behavioural metric trends across difficulty levels.}
    (a)~Source-node convergence speed (lower is better): the average number of attempts to correctly resolve a source node.
    (b)~Premature invocation rate (lower is better): the fraction of non-source invocations made before all predecessors are resolved.
    (c)~Clue adherence rate (higher is better): the fraction of downstream invocations whose arguments trace back to actual upstream outputs.
    All models degrade with difficulty, but at markedly different rates.
  }
  \label{fig:trend}
\end{figure}

We compute three metrics that probe distinct capability dimensions: feedback-driven convergence, reasoning and state management, and environmental clue utilisation.
Figure~\ref{fig:trend} shows the trend of each metric across difficulty levels.

\paragraph{Feedback-driven convergence: source-node convergence speed.}
The \emph{source-node convergence speed} measures how many tool-call attempts are required before a source node (a leaf node with no data-flow predecessors) is correctly resolved.
We focus on source nodes for two reasons.
First, their parameter values are entirely derivable from information visible in the initial environment, removing any confound from upstream computation errors.
Second, the evaluation environment provides explicit per-parameter feedback on each failed attempt---indicating which parameters are correct and which are not---giving the agent a concrete signal to narrow the search space.
This metric therefore directly measures how effectively a model exploits structured feedback to converge on the correct parameter combination: a capable model should use each round of feedback to eliminate incorrect candidates and rapidly zero in on the golden values, while a less capable model will fail to leverage the same information and require many more attempts.

At difficulty-5, efficient models resolve source nodes in fewer than two attempts, whereas GPT-5 requires nearly five---a 3$\times$ gap despite receiving identical feedback.
At difficulty-20, attempt counts rise across all models (range: 2.42--6.52).
The increasing attempt counts at higher difficulties reflect the growing number of source nodes and the correspondingly larger candidate parameter space.
The large spread across models---even when all receive identical per-parameter feedback---reveals that feedback-driven convergence is a genuine differentiating capability rather than a uniform bottleneck.

\paragraph{Reasoning and state management: premature invocation rate.}
The \emph{premature invocation rate} measures the fraction of non-source node invocations that occur before all data-flow predecessors have been resolved---a direct indicator of out-of-order execution and faulty state tracking.
A high rate means the agent attempts to use downstream tools before obtaining the required upstream outputs, a pattern that inevitably leads to incorrect parameter values and wasted steps.
This metric probes a combination of reasoning ability (inferring the correct execution order from environmental clues) and working memory (maintaining an accurate record of which nodes have been solved and which remain pending as the episode progresses).

At difficulty-5, most models maintain rates below 9\%, demonstrating competent topological ordering on shallow graphs.
As difficulty grows, rates increase monotonically for every model, but with markedly different slopes.
At difficulty-20, weaker models (Doubao-Seed-2.0-Pro: 45.1\%; MiniMax-M2: 35.9\%) invoke downstream nodes prematurely in over a third of their calls, while stronger models (Claude-Opus-4.6: 17.1\%; Gemini-3.1-Pro-Preview: 22.1\%) perform comparatively better, though still far from near-zero.
The universal degradation confirms that state management across long dependency chains is a fundamental bottleneck, one that scales superlinearly with graph depth.

\paragraph{Environmental clue utilisation: clue adherence rate.}
The \emph{clue adherence rate} quantifies whether, when a model invokes a node with data-flow predecessors, the argument values it supplies can be traced back to actual outputs of those predecessors.
Low adherence indicates that the model either substitutes self-generated or guessed values, or mistakenly uses outputs from non-dependent nodes, rather than correctly propagating the outputs of its actual upstream predecessors.

At difficulty-5, most models achieve 77--90\% adherence, but by difficulty-20 a sharp separation emerges: Gemini-3.1-Pro-Preview (52.6\%) and Claude-Opus-4.6 (48.0\%) retain substantially higher adherence, while weaker models drop below 20\%, essentially reverting to parametric guessing.
This persistent advantage is the single strongest correlate of leading positions in the main evaluation, suggesting that the ability to faithfully propagate intermediate results through a multi-step tool chain is the core differentiating capability exposed by AgentEscapeBench.

\subsection{Tool-Calling Error Analysis}
\label{sec:error_taxonomy}

We classify every failed tool invocation into a set of error categories by matching the structured feedback returned by the evaluation server (full taxonomy with definitions in Appendix~\ref{app:error_taxonomy}).
Table~\ref{tab:error_dist} shows the distribution across all four difficulty levels.

\begin{table}[t]
  \caption{%
    \textbf{Error type distribution across difficulty levels} (average counts per instance).
    MiniMax-M2 and GPT-5 dominate the missing required parameter category at higher difficulties; GPT-5 uniquely exhibits wrong format errors.
    Error profiles are model-specific, pointing to distinct failure signatures.
  }
  \label{tab:error_dist}
  \centering
  \resizebox{\linewidth}{!}{%
  \begin{tabular}{l*{4}{c}c*{4}{c}c*{4}{c}}
    \toprule
    \rowcolor{HeaderBlue}
    & \multicolumn{4}{c}{\textbf{Miss.\ Required Param.}} && \multicolumn{4}{c}{\textbf{Wrong Node Type}} && \multicolumn{4}{c}{\textbf{Wrong Call Format}} \\
    \cmidrule{2-5} \cmidrule{7-10} \cmidrule{12-15}
    \rowcolor{SubHeaderBlue}
    \textbf{Model} & \textbf{D-5} & \textbf{D-10} & \textbf{D-15} & \textbf{D-20} && \textbf{D-5} & \textbf{D-10} & \textbf{D-15} & \textbf{D-20} && \textbf{D-5} & \textbf{D-10} & \textbf{D-15} & \textbf{D-20} \\
    \midrule
    Claude-Opus-4.6         & 0.0 & 0.0 & 0.1 & 0.3  && 0.2 & 0.3 & 0.3 & 0.3  && 0.0 & 0.0 & 0.0 & 0.0 \\
    \rowcolor{RowGray}
    Gemini-3.1-Pro-Preview  & 0.1 & 0.1 & 0.1 & 0.5  && 0.0 & 0.0 & 0.0 & 0.1  && 0.0 & 0.0 & 0.0 & 0.0 \\
    GPT-5.4                 & 0.2 & 0.7 & 0.7 & 1.1  && 0.0 & 0.1 & 0.0 & 0.1  && 0.0 & 0.0 & 0.0 & 0.0 \\
    \rowcolor{RowGray}
    Kimi-K2.5               & 0.1 & 1.2 & 0.2 & 6.2  && 0.1 & 0.2 & 0.2 & 0.5  && 0.0 & 0.0 & 0.0 & 0.0 \\
    DeepSeek-Chat           & 0.1 & 0.5 & 0.8 & 2.9  && 0.3 & 0.4 & 0.4 & 0.9  && 0.0 & 0.0 & 0.0 & 0.0 \\
    \rowcolor{RowGray}
    DeepSeek-Reasoner       & 0.2 & 0.5 & 0.5 & 1.8  && 0.5 & 0.5 & 0.7 & 0.9  && 0.0 & 0.0 & 0.0 & 0.0 \\
    Gemini-3-Flash-Preview  & 0.0 & 0.1 & 0.2 & 0.5  && 0.0 & 0.0 & 0.1 & 0.1  && 0.0 & 0.0 & 0.0 & 0.0 \\
    \rowcolor{RowGray}
    Doubao-Seed-2.0-Pro     & 0.1 & 0.5 & 0.2 & 1.3  && 0.1 & 0.3 & 0.3 & 0.4  && 0.0 & 0.0 & 0.0 & 0.0 \\
    \rowcolor{StrongRow}
    GPT-5                   & 3.1 & 8.7 & 7.6 & 10.9 && 0.4 & 0.6 & 0.6 & 0.7  && 0.5 & 1.0 & 1.5 & 2.0 \\
    \rowcolor{StrongRow}
    MiniMax-M2              & 0.2 & 2.6 & 7.5 & 13.4 && 0.3 & 0.2 & 0.3 & 0.2  && 0.0 & 0.0 & 0.0 & 0.0 \\
    Kimi-K2                 & 0.1 & 0.8 & 0.8 & 1.9  && 0.1 & 0.1 & 0.2 & 0.1  && 0.0 & 0.0 & 0.0 & 0.0 \\
    \bottomrule
  \end{tabular}%
  }
\end{table}

Three error types reveal distinctive model-specific deficiencies:
\begin{itemize}[leftmargin=1.5em, itemsep=2pt, parsep=0pt, topsep=3pt]
  \item \textbf{Missing required parameter.}
  MiniMax-M2 (13.4 per instance at difficulty-20) and GPT-5 (8.7 at difficulty-10) exhibit dramatically high rates, suggesting systematic failure to enumerate the complete parameter set as tool schemas grow complex.
  \item \textbf{Wrong format.}
  GPT-5 uniquely suffers from this error across difficulty 5--15 while all other models produce near-zero counts, indicating a systematic format adherence failure for unfamiliar APIs.
  \item \textbf{Wrong node type.}
  DeepSeek-Reasoner and DeepSeek-Chat show the highest rates at difficulty-20, reflecting a failure to correctly identify which nodes are executable tools.
\end{itemize}
\noindent
The distinct error profiles confirm that different models exhibit qualitatively different failure signatures, suggesting that targeted improvements could yield disproportionate gains.
\vspace{-0.8\baselineskip}
\subsection{Efficiency Analysis}
\label{sec:efficiency}
\vspace{-0.3\baselineskip}

Beyond correctness, we examine the \emph{average number of tool calls} consumed per instance as a measure of problem-solving efficiency.
Fewer tool calls indicate that the model navigates the dependency graph with less redundancy---fewer incorrect tool invocations and tighter execution plans.
Table~\ref{tab:steps} reports the mean tool-call count for each model--difficulty pair.

\noindent
\begin{minipage}[t]{0.54\linewidth}
  \centering
  \captionof{table}{%
    \textbf{Average tool calls per instance} across difficulty levels.
    Lower is better.
    \textbf{Bold} = fewest in column; \underline{underline} = second fewest.
    $^\star$Min.\ Required: the average minimum number of tool calls needed to solve an instance (theoretical lower bound).
  }
  \label{tab:steps}
  \resizebox{\linewidth}{!}{%
  \begin{tabular}{lcccc}
    \toprule
    \textbf{Model} & \textbf{Diff-5} & \textbf{Diff-10} & \textbf{Diff-15} & \textbf{Diff-20} \\
    \midrule
    \rowcolor{humanrow}
    Min.\ Required$^\star$ & 9.03 & 18.20 & 27.73 & 36.63 \\
    \midrule
    \rowcolor{gray!8}
    Claude-Opus-4.6     & \textbf{13.1} & \underline{36.0} & \textbf{58.7} & \textbf{101.2} \\
    Gemini-3.1-Pro-Preview      & 14.5 & \textbf{35.1} & \underline{66.1} & 150.6 \\
    \rowcolor{gray!8}
    GPT-5.4             & \underline{13.2} & 52.2 & 120.0 & 213.6 \\
    Kimi-K2.5           & 13.7 & 42.0 & 82.6 & 137.6 \\
    \rowcolor{gray!8}
    DeepSeek-Chat       & 15.6 & 39.4 & 78.6 & 130.9 \\
    DeepSeek-Reasoner   & 16.6 & 47.2 & 84.9 & 126.9 \\
    \rowcolor{gray!8}
    Gemini-3-Flash-Preview      & 14.6 & 46.7 & 87.6 & 152.2 \\
    Doubao-Seed-2.0-Pro & 15.4 & 51.4 & 107.7 & 146.9 \\
    \rowcolor{gray!8}
    GPT-5               & 24.8 & 61.5 & 108.4 & \underline{120.4} \\
    MiniMax-M2          & 20.3 & 58.4 & 116.1 & 151.2 \\
    \rowcolor{gray!8}
    Kimi-K2             & 14.2 & 50.5 & 104.2 & 140.9 \\
    \bottomrule
  \end{tabular}%
  }
\end{minipage}%
\hfill
\begin{minipage}[t]{0.43\linewidth}
  \centering
  \vspace{0pt}
  \includegraphics[width=\linewidth]{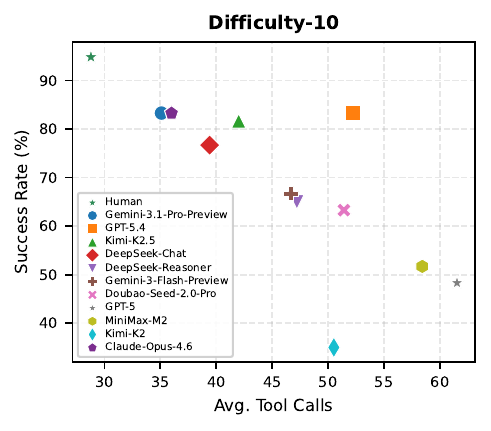}
  \captionof{figure}{%
    \textbf{Tool calls vs.\ success rate} at difficulty-10. Upper-left is better (fewer calls, higher SR).
  }
  \label{fig:efficiency_scatter}
\end{minipage}

From Table~\ref{tab:steps}, efficiency differences \emph{amplify} with difficulty.
At difficulty-5, all models complete within 13--25 tool calls (a modest 1.9$\times$ spread); by difficulty-20, the range expands to 101--214 calls.
Claude-Opus-4.6 and Gemini-3.1-Pro-Preview consistently consume the fewest calls across all levels, staying closest to the theoretical minimum, while GPT-5 and MiniMax-M2 require substantially more.
Notably, GPT-5's low tool-call count at difficulty-20 (120.4, the fewest) does not reflect efficient problem-solving---its low success rate (20.0\%) suggests it terminates episodes prematurely rather than persisting through the full dependency chain.

Figure~\ref{fig:efficiency_scatter} visualises the relationship between efficiency and success rate at difficulty-10, with the human baseline included for reference.
The human data point (upper-left region) demonstrates that humans solve the task with both high accuracy and few tool calls, representing the ideal of directed problem-solving.
Among models, a clear positive correlation emerges: models in the upper-left quadrant (fewer calls, higher SR) such as Claude-Opus-4.6 and Gemini-3.1-Pro-Preview reach solutions via more direct execution paths, whereas models in the lower-right (more calls, lower SR) such as GPT-5 and MiniMax-M2 expend many calls unproductively.
This correlation confirms that AgentEscapeBench rewards \emph{directed} problem-solving rather than exhaustive search.

%% file: related_work.tex
\section{Related Work}
\label{sec:related_work}

Table~\ref{table:related_work} summarizes how AgentEscapeBench compares with representative agent and puzzle-based benchmarks. Existing agent benchmarks often support real tool execution and automatic evaluation, but typically operate in familiar domains with relatively fixed workflows; puzzle-based benchmarks better stress out-of-domain exploration, but often lack executable tools and fine-grained diagnostics. AgentEscapeBench bridges these directions by combining unfamiliar escape-room-style tasks with real tool execution, controlled dependency depth, deterministic evaluation, and step-level trajectory diagnostics.

\begin{table*}[t]
\centering
\setlength{\tabcolsep}{3pt}
\caption{Comparison of representative \emph{agent benchmarks} and \emph{puzzle-based benchmarks}. The table indicates whether each trait is fully addressed ({\cmarkG}), partially addressed ({\notcheckmark}), or not addressed ({\xmarkR}). }
\resizebox{\textwidth}{!}{%
\begin{tabular}{lccccccccc}
\toprule
\multirow{2}{*}{\textbf{Benchmark}} &
\multicolumn{2}{c}{\textbf{Generalization}} &
\multicolumn{3}{c}{\textbf{Tool-use \& Dependency Reasoning}} &
\multicolumn{3}{c}{\textbf{Evaluation \& Diagnostics}} &
\multicolumn{1}{c}{\textbf{Scalability}} \\
\cmidrule(lr){2-3}\cmidrule(lr){4-6}\cmidrule(lr){7-9}\cmidrule(lr){10-10}
& \makecell{OOD\\Setting}
& \makecell{No Fixed\\Workflow}
& \makecell{Real Tool\\Execution}
& \makecell{Deep Contextual\\Dependency}
& \makecell{Minimal Prior\\Knowledge}
& \makecell{Deterministic\\Auto Evaluation}
& \makecell{Step-level\\Diagnostics}
& \makecell{Human Eval.\\\& Comparison}
& \makecell{Controlled\\Difficulty} \\
\midrule

BFCL \citep{patil2025berkeley}
& \xmarkR & \notcheckmark & \cmarkG & \xmarkR & \xmarkR & \cmarkG & \xmarkR & \xmarkR & \notcheckmark \\

$\tau^2$-Bench \citep{barres2025tau}
& \xmarkR & \notcheckmark & \cmarkG & \xmarkR & \notcheckmark & \cmarkG & \notcheckmark & \xmarkR & \xmarkR \\

TripBench \citep{shen2026trip}
& \xmarkR & \notcheckmark & \cmarkG & \notcheckmark & \notcheckmark & \cmarkG & \notcheckmark & \xmarkR & \cmarkG \\

VitaBench \citep{he2025vitabench}
& \xmarkR & \notcheckmark & \cmarkG & \notcheckmark & \notcheckmark & \xmarkR & \notcheckmark & \xmarkR & \notcheckmark \\

SWE-bench \citep{jimenez2023swe}
& \xmarkR & \notcheckmark & \cmarkG & \xmarkR & \xmarkR & \cmarkG & \xmarkR & \xmarkR & \xmarkR \\

EscapeBench \citep{qian2024escapebench}
& \cmarkG & \cmarkG & \xmarkR & \notcheckmark & \notcheckmark & \notcheckmark & \xmarkR & \xmarkR & \notcheckmark \\

PuzzleWorld \citep{li2025puzzleworld}
& \cmarkG & \cmarkG & \xmarkR & \notcheckmark & \notcheckmark & \notcheckmark & \xmarkR & \xmarkR & \notcheckmark \\

MCP-Bench \citep{wang2025mcp}
& \notcheckmark & \notcheckmark & \cmarkG & \xmarkR & \xmarkR & \xmarkR & \cmarkG & \xmarkR & \xmarkR \\

\midrule
\textbf{AgentEscapeBench (ours)}
& \cmarkG & \cmarkG & \cmarkG & \cmarkG & \cmarkG & \cmarkG & \cmarkG & \cmarkG & \cmarkG \\
\bottomrule
\end{tabular}%
}
\label{table:related_work}
\vspace{-6pt}
\end{table*}

\textbf{Agent Benchmarks with Domain-Specific Tasks.}
A large body of agent benchmarks focuses on tasks drawn from well-defined, real-world domains where models carry substantial prior knowledge.
Representative examples include software engineering~\citep{jimenez2023swe}, travel planning~\citep{shen2026trip}, retail and telecom services~\citep{barres2025tau}, smartphone applications~\citep{russell2025gaia}, tool-call accuracy~\citep{patil2025berkeley}, code and web navigation~\citep{liu2023agentbench}, real-world applications~\citep{he2025vitabench}, and long-horizon task completion~\citep{luo2025ultrahorizon}.
While these benchmarks measure important capabilities, their domain-specific nature means that a model can succeed largely through prior knowledge of the task workflow.
AgentEscapeBench intentionally provides \emph{no} domain prior: the agent must discover solution paths through evidence-based inference in an unfamiliar environment, targeting out-of-domain generalisation rather than in-domain task mastery.

\textbf{Escape-Room and Puzzle-Based Benchmarks.}
Several recent works share our insight that open-ended, unstructured tasks better probe exploratory reasoning.
EscapeBench~\citep{qian2024escapebench} adopted text-based room-escape scenarios; VisEscape~\citep{lim2025visescape} leveraged a visual room-escape framing; PuzzleWorld~\citep{li2025puzzleworld} used multi-step puzzle-hunts; BigEscape~\citep{tang2025big} drew from television escape shows.
However, all of these benchmarks either restrict the agent to predefined text actions or rely on pure question answering, without invoking real external APIs.
AgentEscapeBench addresses this gap by coupling open-ended exploration with genuine function calling, making the evaluated capabilities directly transferable to real-world agentic deployments.

\textbf{Reasoning Benchmarks.}
Mathematical reasoning benchmarks such as MATH~\citep{hendrycks2021measuring} and AIME~\citep{AIME25} evaluate pure reasoning ability without tool interaction, while \citet{zhang2023evaluating} specifically studied tool-augmented math reasoning, demonstrating that tool use extends model competency into domains requiring deep abstract computation.
More broadly, \citet{qiao2024making} showed that real tool use pushes past the capability boundaries of pure text-based inference.
AgentEscapeBench complements these efforts by evaluating reasoning that is inherently \emph{grounded} in tool interactions: rather than using tools to assist known problem types, agents must reason about unfamiliar tools and adaptively chain their outputs---a capability that existing reasoning benchmarks do not measure.

%% file: conclusion.tex
\section{Conclusion}
\label{sec:conclusion}

We presented \textbf{AgentEscapeBench}, a benchmark that evaluates general tool-grounded reasoning in deliberately unfamiliar problem settings.
Tasks are constructed as directed acyclic graphs with no fixed solution pattern, requiring genuine function calls and adaptive planning under incremental information disclosure.
Our evaluation of sixteen LLMs reveals that the benchmark effectively differentiates models across a wide capability spectrum, with performance gaps amplifying sharply at higher difficulties, and that fine-grained trajectory analysis exposes model-specific failure signatures that provide actionable directions for improving general-purpose agent capabilities.

%% file: appendix_data_construction.tex
\section{Dataset Construction Details}
\label{app:data_construction}

This appendix provides full algorithmic details for the six-stage data construction pipeline summarised in Section~\ref{sec:data_construction}.

\subsection{Stage 1: Tool and Item Template Library}

The template library constitutes the atomic vocabulary from which every puzzle instance is assembled.
It comprises four categories:

\begin{itemize}[leftmargin=1.5em]
  \item \textbf{Tool templates (32).}
  Each template defines typed input ports, typed output ports, and a deterministic execution function.
  The library covers:
  big-integer arithmetic (multiplication, modular exponentiation, GCD, modular inverse),
  cryptographic primitives (SHA-256, MD5, HMAC-SHA256, AES-CBC decryption, RSA decryption),
  encoding/decoding (Base64, Hex, Zlib, ROT-N cipher, XOR),
  file and archive operations (ZIP extraction, CSV query, JSON path query, regex search),
  graph algorithms (shortest path),
  code execution (Python executor),
  numeric conversion (base converter, CRC32, Luhn check, IBAN validation),
  and text processing (Unicode normalisation, Bidi sanitisation, timezone conversion).

  \item \textbf{Terminal templates (2).}
  Special tool nodes that serve as the execution endpoint.

  \item \textbf{Item templates (16).}
  Information-carrying entities that may hold textual payloads, numerical values, or file references.

  \item \textbf{Container templates (4).}
  Entities with \emph{drop mechanisms}: interacting with a container reveals hidden items or tools that were initially invisible.
\end{itemize}

Port types are drawn from a closed semantic vocabulary:
\texttt{Big\_Int}, \texttt{Text\_Generic}, \texttt{Hex\_String}, \texttt{Hex\_String\_Key\_AES}, \texttt{Hex\_String\_IV\_AES}, \texttt{File\_Id}, \texttt{Code\_Python}, \texttt{Item}, and \texttt{Hidden\_Item}.

In addition, 68 \emph{droppable tool variants} are automatically derived from the base tool templates by adding \texttt{ITEM}/\texttt{HIDDEN\_ITEM} trigger ports.
These variants are mixed into the generation pool with a configurable ratio (default 30\%), enabling the benchmark to include tools that are only revealed through the incremental-disclosure mechanism.

\subsection{Stage 2: DAG Skeleton Generation (Reverse-Generation Algorithm)}

\begin{algorithm}
\caption{DAG Skeleton Generation via Reverse Growth}
\label{alg:dag_generation}
\begin{algorithmic}[1]
\Require Target node count $n$, template library $\mathcal{T}$
\Ensure DAG skeleton $G = (V, E)$
\State Select a random template $t \in \mathcal{T}$ that has $\geq 1$ input port
\State Create goal node $v_{\mathrm{goal}}$ from $t$; set $V \leftarrow \{v_{\mathrm{goal}}\}$, $E \leftarrow \emptyset$
\State Initialize pending queue $Q$ with all input ports of $v_{\mathrm{goal}}$
\While{$|V| < n$ and $Q \neq \emptyset$}
    \State Pop a random pending requirement $(v, p, \tau, r)$ from $Q$
    \If{$r > 5$}
        \State \textbf{continue} \Comment{skip unresolvable requirement}
    \EndIf
    \State Compute reuse probability $p_r \leftarrow 0.3 + 0.3 \cdot |V| / n$
    \State Find candidates $\mathcal{C} \leftarrow \{u \in V \mid u$ outputs type $\tau$, $u \neq v$, no path $v \leadsto u\}$
    \If{$\mathcal{C} \neq \emptyset$ \textbf{and} $\mathrm{rand}() < p_r$}
        \State Pick $u \in \mathcal{C}$ at random; select available output port $q$ of type $\tau$
        \State Add edge $(u, q) \rightarrow (v, p)$ to $E$
    \Else
        \State Gather candidate templates from $\mathcal{T}$ with an output of type $\tau$
        \If{no candidates exist}
            \State Re-enqueue $(v, p, \tau, r{+}1)$ into $Q$; \textbf{continue}
        \EndIf
        \State Sample template $t'$; create new node $u$ from $t'$
        \State $V \leftarrow V \cup \{u\}$; add edge $(u, q) \rightarrow (v, p)$ to $E$
        \State Enqueue all input ports of $u$ into $Q$ with $r = 0$
    \EndIf
\EndWhile
\If{$G$ is isomorphic to a previously accepted skeleton}
    \State Discard $G$ and restart from line 1
\EndIf
\State \Return $G$
\end{algorithmic}
\end{algorithm}

Given a target node count $n$ (the difficulty level), the algorithm proceeds as follows:

\begin{enumerate}[leftmargin=1.5em]
  \item \textbf{Sample final-goal node.}
  A template with at least one input port is selected uniformly at random.
  A node is instantiated from this template and designated as the win node; the puzzle's success condition is defined as producing the correct output of this node.

  \item \textbf{Initialise pending queue.}
  All input ports of the final-goal node are enqueued as pending requirements, each represented as a tuple $(v, p, \tau, r)$: target node $v$, target port $p$, required type $\tau$, and retry count $r=0$.

  \item \textbf{Iterative resolution.}
  While $|\text{nodes}| < n$ and the queue is non-empty:
  \begin{enumerate}
    \item Pop a random pending requirement $(v, p, \tau, r)$; skip if $r > 5$.
    \item With probability $0.3 + 0.3 \cdot (|\text{nodes}| / n)$, attempt to \emph{reuse} an existing node whose output type matches $\tau$, subject to acyclicity and port-availability checks.
    \item Otherwise, instantiate a \emph{fresh} template node whose output matches $\tau$, add it to the graph, create the connecting edge, and enqueue all input ports of the new node.
  \end{enumerate}

  \item \textbf{Structural constraints} enforced at every edge creation:
  \begin{itemize}
    \item \emph{Acyclicity}: BFS reachability check vetoes any edge that would close a cycle.
    \item \emph{Physical-object single-use}: ports of type \texttt{ITEM}/\texttt{HIDDEN\_ITEM} may appear on at most one edge.
    \item \emph{Special-port exclusivity}: semantically sensitive input ports (regex patterns, JSON paths, OTP secrets, Base64 data) consume their source output exclusively---once connected, the source port is barred from further reuse.
    \item \emph{Same-target deduplication}: a target node's distinct input ports may not draw from the same source port.
  \end{itemize}
\end{enumerate}

\paragraph{Isomorphism checking.}
Each newly generated skeleton is checked against all previously accepted ones.
Two DAGs are declared isomorphic if they share the same multiset of template identifiers, the same multiset of edge patterns (source template, source port, target template, target port), and the same total node count.
Isomorphic duplicates are discarded.

\subsection{Stage 3: Source Annotation}

The annotation pass identifies every port that requires a concrete seed value before the forward execution pass.
Two categories are flagged:

\begin{enumerate}[leftmargin=1.5em]
  \item \textbf{Unconnected input ports}: any input port with no incoming edge is a leaf-level source.
  \item \textbf{Drop-node outputs}: if a node's triggering input is of type \texttt{ITEM}/\texttt{HIDDEN\_ITEM} (indicating activation by a physical object from a container), its output ports that feed downstream nodes are also flagged, because these values must be materialised before forward execution can propagate through them.
\end{enumerate}

The output is a structured \texttt{source\_init} map attached to the DAG metadata.

\subsection{Stage 4: Value Instantiation via LLM}

For each port in the \texttt{source\_init} map, an LLM (accessed via an OpenAI-compatible API) generates a concrete seed value.
Each request provides the port's semantic type, its natural-language description, and any inter-port consistency constraints---for example:
\begin{itemize}[leftmargin=1.5em]
  \item A modulus must exceed its base in a modular-exponentiation chain.
  \item A file path must be syntactically valid for the target operation.
  \item An AES key must be exactly 16/24/32 bytes in hex encoding.
\end{itemize}

This step introduces lexical and semantic diversity while ensuring every value satisfies its downstream type contract.

\subsection{Stage 5: Deterministic Forward Execution}

Once all source ports carry concrete values, the DAG is executed in topological order by a deterministic simulator.
Each tool node reads the values on its input ports, applies its pre-defined computation function, and writes results to its output ports.
Values are immediately propagated along outgoing edges.
Execution continues until every port holds a concrete value, yielding a fully instantiated DAG whose terminal output is the unique ground-truth flag.

A post-execution \textbf{completeness check} verifies that all non-physical-object ports have been assigned a value.
Instances that fail this check are moved to a quarantine directory for post-hoc analysis.

\subsection{Stage 6: Narrative Generation and Quality Filtering}

\paragraph{Narrative generation.}
Each node receives a background narrative generated by an LLM, conditioned on its DAG connectivity.
The prompt supplies the node's type, its upstream and downstream relationships, and the thematic style.
Narratives partially disclose the data-flow structure---hinting at which upstream output feeds a given downstream input---without revealing the complete solution path.

To promote stylistic diversity and mitigate exploitation of surface-level textual cues, each instance is randomly assigned one of eight thematic styles:
\textit{cyberpunk}, \textit{detective}, \textit{sci-fi}, \textit{fantasy}, \textit{steampunk}, \textit{post-apocalyptic}, \textit{mystery}, and \textit{techno-thriller}.

\paragraph{Final validation.}
Before admission to the benchmark pool, each instance passes automated checks:
\begin{itemize}[leftmargin=1.5em]
  \item All required narrative fields are present.
  \item All node references are resolvable within the DAG.
  \item Every non-physical-object input port has a value; every non-physical-object output port has a value.
  \item The win condition is reachable via at least one complete tool-call sequence.
\end{itemize}
Instances that fail any check are moved to a rejection directory, guaranteeing that every accepted instance is well-formed, solvable, and has a provably correct unique answer.

%% file: appendix_evaluation.tex
\section{Evaluation Details}
\label{app:evaluation}

\paragraph{Step budget.}
To accommodate the increasing number of tool calls required at higher difficulty levels, we set the maximum number of agent actions to 35, 80, 130, 160, and 200 for difficulty levels 5, 10, 15, 20, and 25, respectively.
An episode terminates when the agent either submits the correct flag, exhausts its step budget, or explicitly declares failure.

\paragraph{Model configuration.}
All models are evaluated via their official API endpoints with a sampling temperature of 0.7.
Each instance is evaluated in a single run; we do not average over multiple trials.
We evaluate sixteen models spanning frontier closed-source systems (Claude-Opus-4.6, GPT-5.4, GPT-5, Gemini-3.1-Pro-Preview, Gemini-3-Flash-Preview, Kimi-K2.5, Kimi-K2, DeepSeek-Chat, DeepSeek-Reasoner, Doubao-Seed-2.0-Pro, MiniMax-M2) and open-source alternatives (Qwen3-235B-A22B, Qwen3-Next-80B-A3B, Qwen3-32B, Qwen3-14B, Qwen3-8B).

\subsection{Error Type Taxonomy}
\label{app:error_taxonomy}

We classify every failed tool invocation (excluding \texttt{investigate} and \texttt{submit\_answer} operations) into the following categories based on the structured feedback returned by the evaluation server.

\begin{enumerate}[leftmargin=1.5em]
  \item \textbf{Missing required parameter} — The agent omits one or more mandatory input parameters when invoking a tool or item.
  Triggered when the server reports missing parameters, an empty parameter set, or a partial parameter submission.

  \item \textbf{Wrong node type} — The agent attempts an operation incompatible with the node's type, such as calling \texttt{use\_item} with arguments on a simple Item that should be accessed via \texttt{investigate}.

  \item \textbf{Repeated solved node} — The agent attempts to unlock an item that has already been successfully solved in a previous step, wasting a turn.

  \item \textbf{Wrong parameter type} — The agent provides a parameter value with an incorrect data type (e.g., passing an integer where a hex string is expected).

  \item \textbf{Node not visible} — The agent attempts to invoke a node that does not currently exist in the environment, or mistakenly treats a non-node entity as an invocable node.

  \item \textbf{Node not exist} — The agent references a node ID that does not exist in the current scenario, typically due to hallucinating a node name or confusing template names with instance-specific IDs.

  \item \textbf{Wrong format} — The agent's tool-call request is structurally malformed, missing essential fields such as \texttt{node\_id}.

  \item \textbf{Other} — Errors that do not match any of the above classification rules.
\end{enumerate}

%% file: appendix_trajectory.tex
\section{Example Solving Trajectory}
\label{app:trajectory}

\begin{figure}[H]
  \centering
  \includegraphics[width=\linewidth]{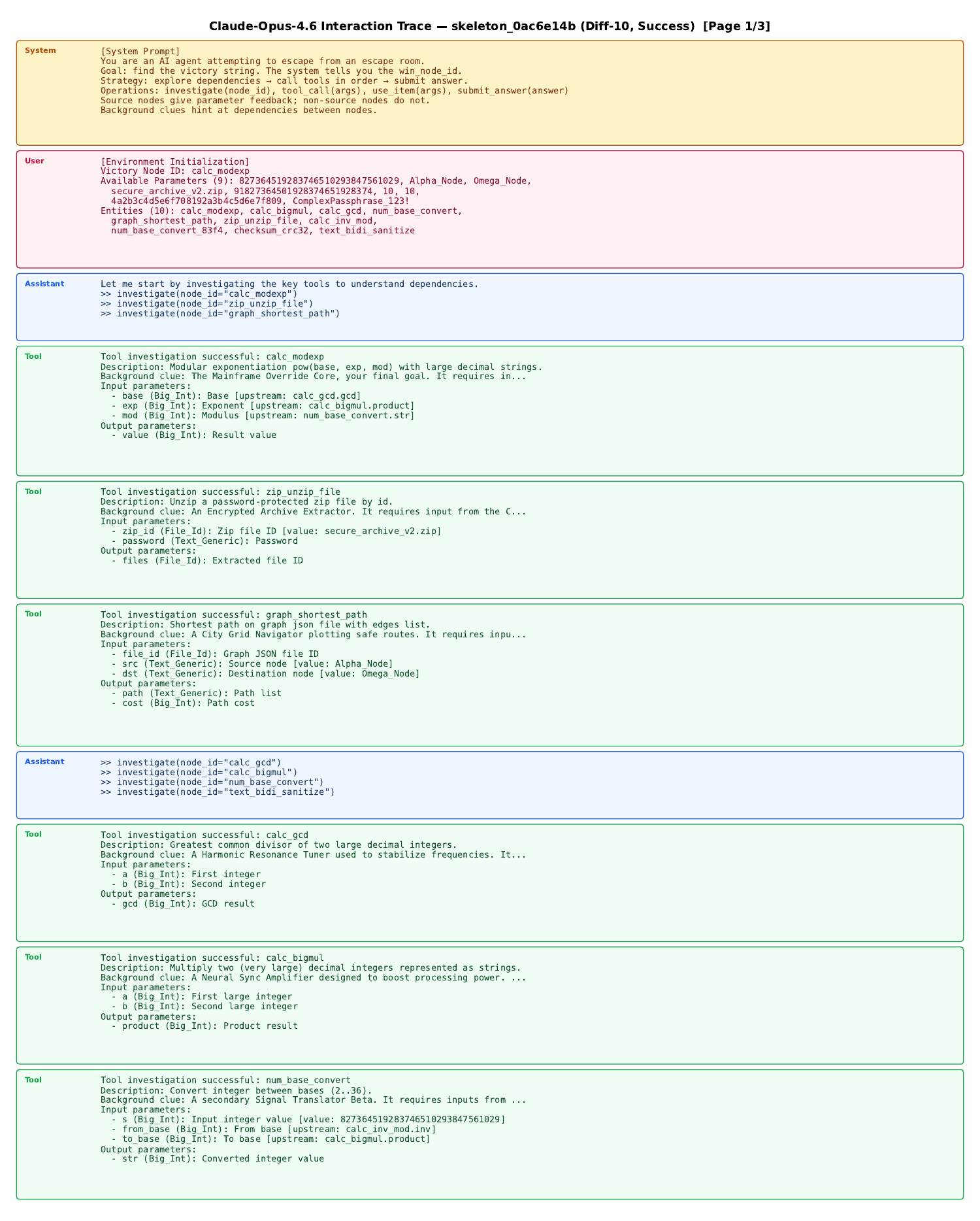}
  \caption{Interaction trace (page 1/3): system prompt, environment initialization, and first investigation round.}
  \label{fig:traj_p1}
\end{figure}

\begin{figure}[H]
  \centering
  \includegraphics[width=\linewidth]{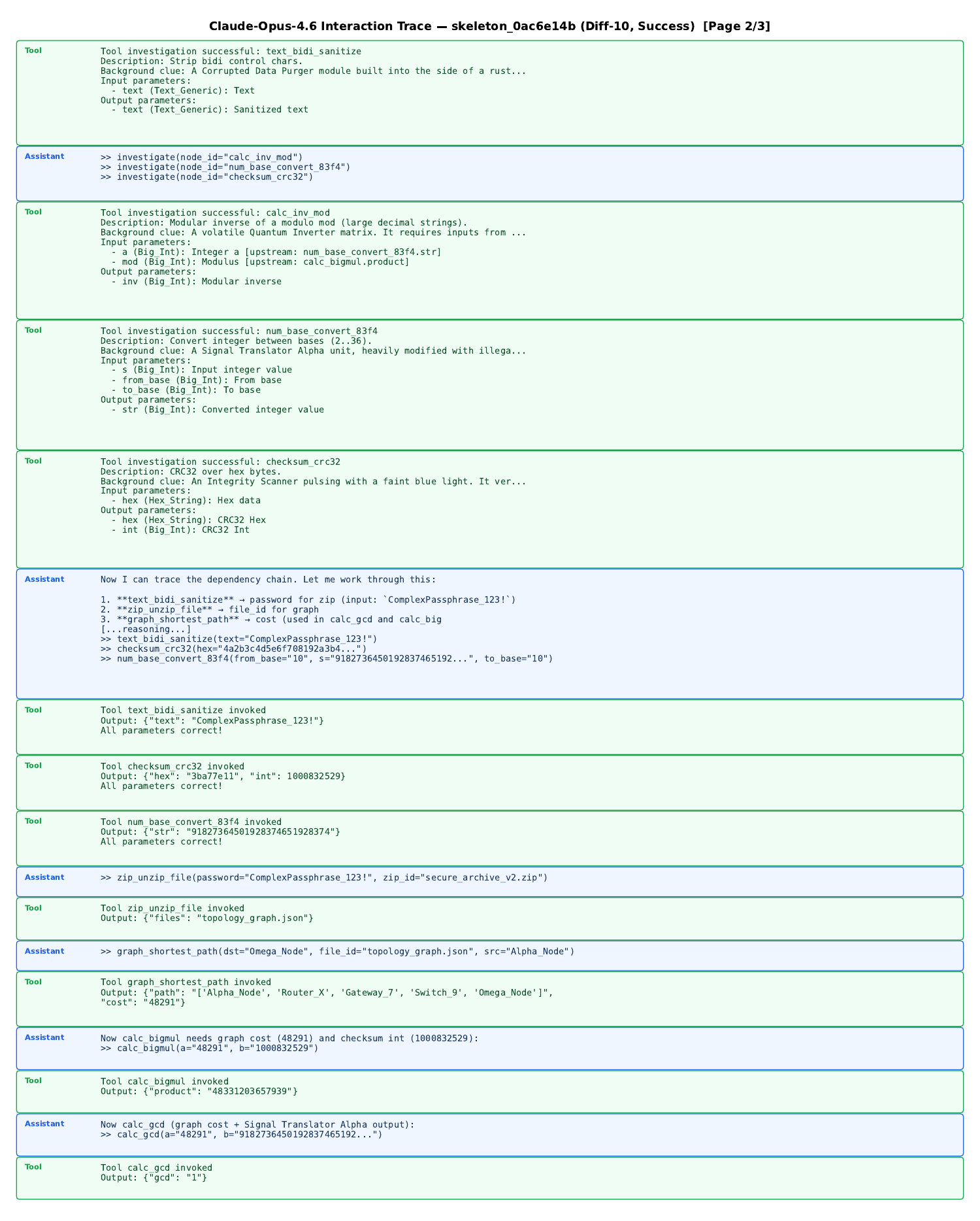}
  \caption{Interaction trace (page 2/3): remaining investigations, dependency reasoning, and tool execution chain.}
  \label{fig:traj_p2}
\end{figure}

\begin{figure}[H]
  \centering
  \includegraphics[width=\linewidth]{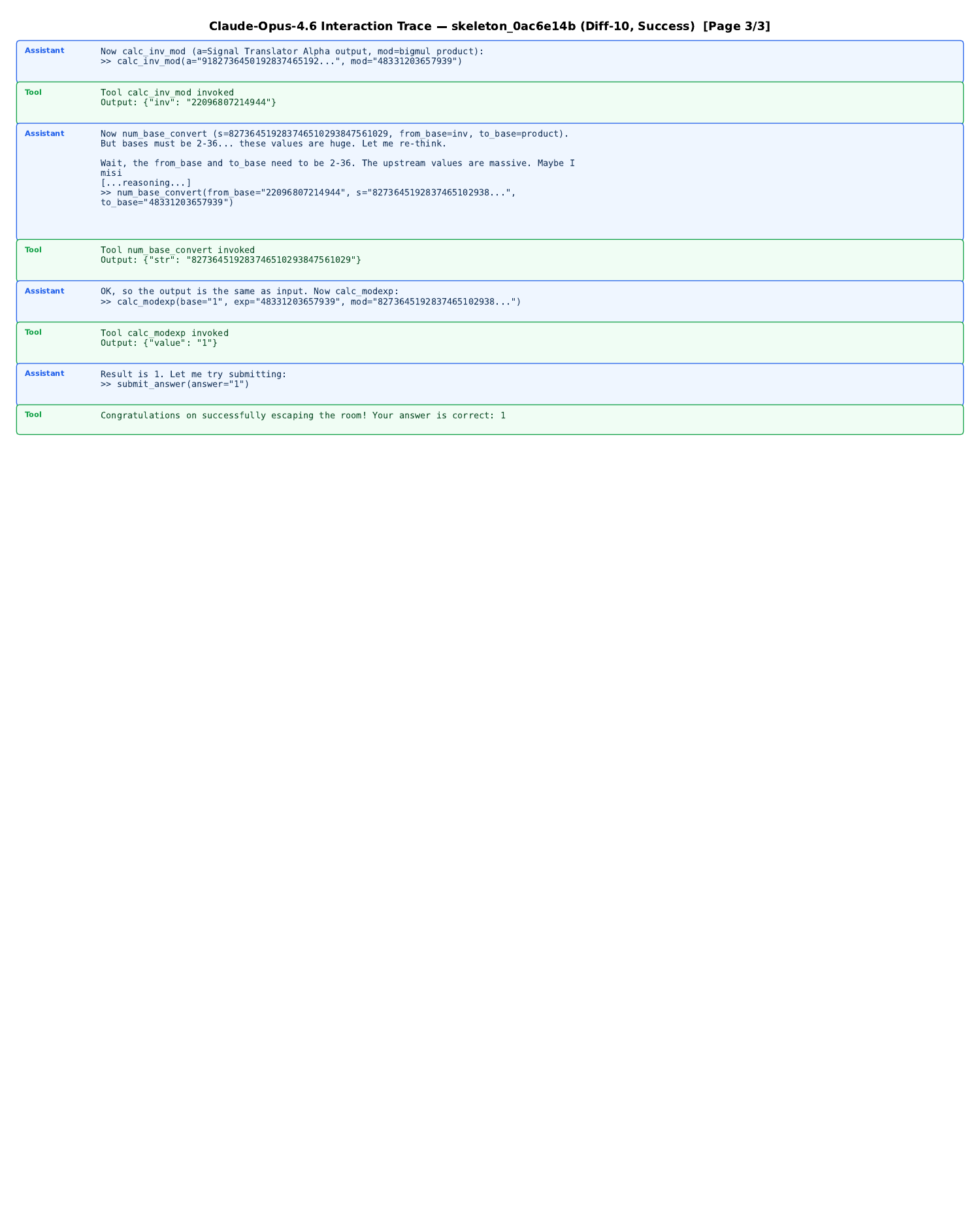}
  \caption{Interaction trace (page 3/3): final computation steps and successful answer submission.}
  \label{fig:traj_p3}
\end{figure}

We present a complete interaction trace of Claude-Opus-4.6 solving a difficulty-10 instance.
The agent successfully solves the puzzle in 12 turns (3 investigation rounds comprising 10 investigations, 8 tool-call rounds comprising 10 tool invocations, and 1 answer submission) with zero errors.
Figures~\ref{fig:traj_p1}--\ref{fig:traj_p3} show the full context visible to the model, including the system prompt, environment initialization, and all agent--environment interactions.

%% file: appendix_limitations.tex
\section{Limitations}
\label{sec:limitations}

\paragraph{Language coverage.}
All narrative descriptions and tool interfaces in AgentEscapeBench are in English.
Evaluating multilingual tool-use reasoning remains future work.

\paragraph{Single-trial evaluation.}
Each instance is evaluated in a single run per model due to the substantial API cost of running sixteen models across 270 instances.
While this reflects practical deployment conditions, it does not capture variance across runs; future work may consider multiple trials with confidence intervals.

\paragraph{Fixed step budgets.}
Step budgets are set per difficulty level and are uniform across models.
Some models may benefit from larger budgets at higher difficulties, and adaptive budgeting strategies remain unexplored.